\documentclass[runningheads]{llncs}
\usepackage{graphicx}
\usepackage{amsmath,amssymb} % define this before the line numbering.
\usepackage{color}
\usepackage{subfigure}
\usepackage{float}
\usepackage{comment}
\usepackage{booktabs}
\usepackage{multirow}
\usepackage{arydshln}
\usepackage[colorlinks,
            linkcolor=blue,
            anchorcolor=blue,
            citecolor=blue]{hyperref}

\begin{document}
\pagestyle{headings}
\mainmatter

\def\ACCV20SubNumber{360}  % Insert your submission number here

%===========================================================
\title{Background Learnable Cascade for Zero-Shot Object Detection} % Replace with your title
\titlerunning{BLC for Zero-Shot Object Detection}
% If the paper title is too long for the running head, you can set
% an abbreviated paper title here
%
\author{Ye Zheng\inst{1,2}\orcidID{0000-0003-1618-6834} \and
Ruoran Huang\inst{1,2}\orcidID{0000-0001-9014-761X} \and
Chuanqi Han\inst{1,2}\orcidID{0000-0002-3482-0475} \and Xi Huang\inst{1}\orcidID{0000-0003-1953-5809} \and Li Cui\inst{1}\orcidID{0000-0002-4125-2138}}
\authorrunning{Y. Zheng et al.}
% First names are abbreviated in the running head.
% If there are more than two authors, 'et al.' is used.
%
\institute{Institute of Computing Technology, Chinese Academy of Sciences, Beijing 100190, China \and
University of Chinese Academy of Sciences, Beijing 100190, China}
% \email{lncs@springer.com}\\
% \url{http://www.springer.com/gp/computer-science/lncs}

\maketitle

%===========================================================
\begin{abstract}
Zero-shot detection (ZSD) is crucial to large-scale object detection with the aim of simultaneously localizing and recognizing unseen objects. There remain several challenges for ZSD, including reducing the ambiguity between background and unseen objects as well as improving the alignment between visual and semantic concept. In this work, we propose a novel framework named Background Learnable Cascade (BLC) to improve ZSD performance. The major contributions for BLC are as follows: (i) we propose a multi-stage cascade structure named Cascade Semantic R-CNN to progressively refine the alignment between visual and semantic of ZSD; (ii) we develop the semantic information flow structure and directly add it between each stage in Cascade Semantic R-CNN to further improve the semantic feature learning; (iii) we propose the background learnable region proposal network (BLRPN) to learn an appropriate word vector for background class and use this learned vector in Cascade Semantic R-CNN, this design makes ``Background Learnable" and reduces the confusion between background and unseen classes. Our extensive experiments show BLC obtains significantly performance improvements for MS-COCO over state-of-the-art methods. $\footnote{Code has been made available at \href{https://github.com/zhengye1995/BLC}{https://github.com/zhengye1995/BLC}}$

\keywords{Zero-shot object detection, Multi-stage structure, Background learnable, Semantic information flow}
\end{abstract}

%===========================================================
\section{Introduction}

% Please follow the steps outlined below when submitting your manuscript.
%Do not use any additional Latex macros.
Zero-shot learning (ZSL) is widely used to reason about objects belonging to unseen classes that have never been observed during training. Traditional ZSL researches focus on the classification problem of unseen objects and achieve high classification accuracy~\cite{zhang2016zero1}. However, there still exists a large gap between ZSL settings and real-world scenarios. ZSL just focuses on recognizing unseen objects, not detecting them. For example, most of datasets used as ZSL benchmark only have one dominant object in each sample~\cite{nilsback2008automated,russakovsky2015imagenet,welinder2010caltech}, while in real-world, various objects may appear in a single image without being precisely localized.

To simultaneously localize and recognize unseen objects, some preliminary attempts~\cite{bansal2018zero,demirel2018zero,rahman2018zero,zhu2019zero} for zero-shot object detection (ZSD) have been reported. ZSD introduces a more practical setting to detect novel objects that are not observed during training. On this foundation, Rahman et al.~\cite{rahman2020improved}, Li et al.~\cite{li2019zero}, Zhao et al.~\cite{zhao2020gtnet} and Zhu et al.~\cite{zhu2020don} make improvements to boost ZSD performance.
These achievements combine the visual-semantic mapping relationship in ZSL with the deep learning based detection model in traditional object detection methods to detect unseen objects. However, these works still have their limitations: (i) can not gradually optimize the visual-semantic alignment to properly map visual features to semantic information; (ii) lack of a handy pipeline to learn a discriminative background class semantic embedding representation, while this representation is important for reducing the confusion between background and unseen classes; (iii) rely on pre-trained weights that were learned from seen or unseen datasets.

We therefore propose a novel framework named Background Learnable Cascade (BLC) for ZSD, including three components: Cascade semantic R-CNN, semantic information flow and BLRPN. BLC is motivated on the cognitive science about how humans reason objects through semantic information. Humans can use semantic information such as words to describe the characteristics of objects, and conversely, humans can also reason the categories for objects from the semantic description. Based on the past life experience, humans have established an abstract visual-semantic mapping relationship for seen objects and transfer it to recognize unseen objects. For example, humans can recognize the zebra with the language description ``a horse with black and white stripes" and the visual memory of horse even if they had never seen a zebra before. Inspirited by this, BLC develops a visual-semantic alignment substructure named semantic branch to learn the visual-semantic relationship between seen objects' images and word vectors. Then transfers this alignment from seen classes to unseen classes to detect unseen objects. In order to progressively refine the visual-semantic alignment, BLC develops Cascade Semantic R-CNN by integrating the semantic branch in a multi-stage architecture based on Cascade R-CNN~\cite{cascadercnn}. This combination can take advantage of the cascade structure and multi-stage refinement policy. In Cascade Semantic R-CNN, the semantic branches in later stages only benefit from better localized bounding boxes without direct semantic information connections. To remedy this problem, BLC further designs semantic information flow structure to improve the semantic information flow by directly connecting semantic branches in each cascade stage. The semantic feature in the current stage will be modulated through fully connected layers and fed to the next stage. This design promotes the circulation of semantic information between each stage and is beneficial to learn a proper visual-semantic relationship. Due to the coarse word vector for background class used in semantic branch is inability to exactly represent the complex background, BLC develops a novel framework denoted as background learnable region proposal network (BLRPN) to learn an appropriate word vector for background class. Our study shows that replacing the coarse background word vector in semantic branch with the new one learned from BLRPN can effectively increases the recall rate for unseen classes.

Our main contributions of Background Learnable Cascade (BLC) are: (i) we develop Cascade Semantic R-CNN, which effectively integrates multi-stage structure and cascade strategy into zero-shot object detection by first integrating cascade with the semantic branch; (ii) we develop semantic information flow structure among each cascade stage to improve the semantic feature learning; (iii) we develop a background learnable region proposal network (BLRPN) to learn a more appropriate background class semantic word vector reducing the confusion between background and unseen classes; (iv) extensive experiments on two different MS-COCO splits show significant performance improvement in terms of mAP and recall.

%------------------------------------------------------------------------- 
\section{Related Work}

\subsubsection{Zero-shot Recognition.}
In the past few years, several works have been proposed~\cite{bendale2016towards,changpinyo2016synthesized,elhoseiny2013write,frome2013devise,jain2014multi,kodirov2017semantic,lampert2009learning,lampert2013attribute,nilsback2008automated,norouzi2013zero,rahman2018unified,xian2017zero,zhang2015zero,zhang2016zero1,zhang2016zero2} for zero-shot image recognition. Most approaches of ZSL~\cite{al2017automatic,al2016recovering,lampert2013attribute,xian2018zero,zablocki2019context,luo2019context,krishna2017visual,mishra2018generative,kumar2018generalized,verma2017simple,verma2020meta} have employed the relationship between seen and unseen classes to optimize recognition of unseen objects. The most classic way is to learn the alignment between the visual and semantic information by using extra source data. This alignment can classify unseen image categories by using labeled image data and semantic representations trained with unsupervised fashion from unannotated text data. In our work, we follow this methodology to detect objects for unseen classes.

%------------------------------------------------------------------------- 
\subsubsection{Object detection.}
Deep learning based object detection methods have made great progress in the past several years, e.g., YOLO~\cite{yolo}, SSD~\cite{ssd}, RetinaNet~\cite{retinaNet}, Faster R-CNN~\cite{fasterrcnn}, R-FCN~\cite{rfcn}, MASK R-CNN~\cite{maskrcnn}, DCN~\cite{dcn}, CornerNet~\cite{cornernet}, CenterNet~\cite{centernet} and FCOS~\cite{fcos}. The recent multi-stage structures have further boosted performance for object detection, e.g., Cascade R-CNN~\cite{cascadercnn} and Cascade RPN~\cite{cai2018cascade}. The multi-stage cascade strategy progressively refine the results and we also adopt this strategy to refine visual-semantic alignment in our BLC.
%------------------------------------------------------------------------- 
\subsubsection{Recent achievements for ZSD.}
In recent years, some ZSD approaches have been proposed. Rahman et al.~\cite{rahman2018zero} combine ConSE~\cite{norouzi2013zero} and Faster R-CNN~\cite{fasterrcnn} with a max-margin loss and a meta-class clustering loss to tackle the problem of ZSD. Bansal et al.~\cite{bansal2018zero} employ a background-aware model to solve the confusion for background class in ZSD, and they use additional data to densely sample training classes. They also propose a generalization version of ZSD called generalized zero-shot object detection (GZSD) which aims to detect seen and unseen objects together. Demirel et al.~\cite{demirel2018zero} adopt the hybrid region embedding to improve performance. Zhu et al.~\cite{zhu2019zero} introduce ZS-YOLO, which is built on a one-step YOLOv2~\cite{redmon2017yolo9000} detector. Rahman et al.~\cite{rahman2020improved} propose polarity loss to cluster semantic and develop an end-to-end network based RetinaNet~\cite{retinaNet}. Li et al.~\cite{li2019zero} address ZSD with textual descriptions by jointly learning visual units, visual-unit attention and word-level attention. 

There are some key differences between our work and previous works: (i) to the choice of evaluation datasets, Rahman et al.~\cite{rahman2018zero} and Zhu et al.~\cite{zhu2019zero} use the ILSVRC-2017 detection dataset~\cite{russakovsky2015imagenet}. This dataset is restrictive for evaluate ZSD, in comparison with our choice --- MS-COCO~\cite{lin2014microsoft}. Because each image in ILSVRC-2017 detection dataset only has one dominant object, which exists a big gap with the real scene. We follow the choices and splits for dataset introduced by Bansal et al.~\cite{bansal2018zero} and Rahman et al.~\cite{rahman2020improved} in MS-COCO). These dataset splits are more challenging and closer to the real scene settings. (ii) for the representation of background class, most of them just use a trivial representation for background class, e.g., the semantic vectors for ‘background’ word~\cite{bansal2018zero} and the mean vectors for all seen classes~\cite{rahman2018zero}. These representations are not the optimal solution to address the confusion between background and unseen classes. Bansal et al.~\cite{bansal2018zero} propose a background-aware approach based on an iterative EM-like training procedure, but it is complex and inefficient for datasets with a small number of categories like MS-COCO. In contrast, our BLRPN, as an end-to-end framework, can learn a reasonable representation for background class through only one training process without iterations while not be affected by the sparsity of category; (iii) in the aspect of the optimization strategy, all of these previous works just refine the visual-semantic alignment once, which may not enough to optimize this alignment. In BLC, we adopt multi-stage architecture to progressively refine this alignment to improve the performance of ZSD. (iv) for the training process, most of them need fine tune their model based on additional pre-trained weights, which are learned from seen or unseen-class data, while our work, as stated above, just needs a simple and straightforward training process without any additional pre-trained weights on seen or unseen data.
%------------------------------------------------------------------------- 
\begin{figure}[tbp]
\centering
\subfigure[Cascade Semantic R-CNN]{
\label{fig:figure1a}
\includegraphics[width=6.5cm]{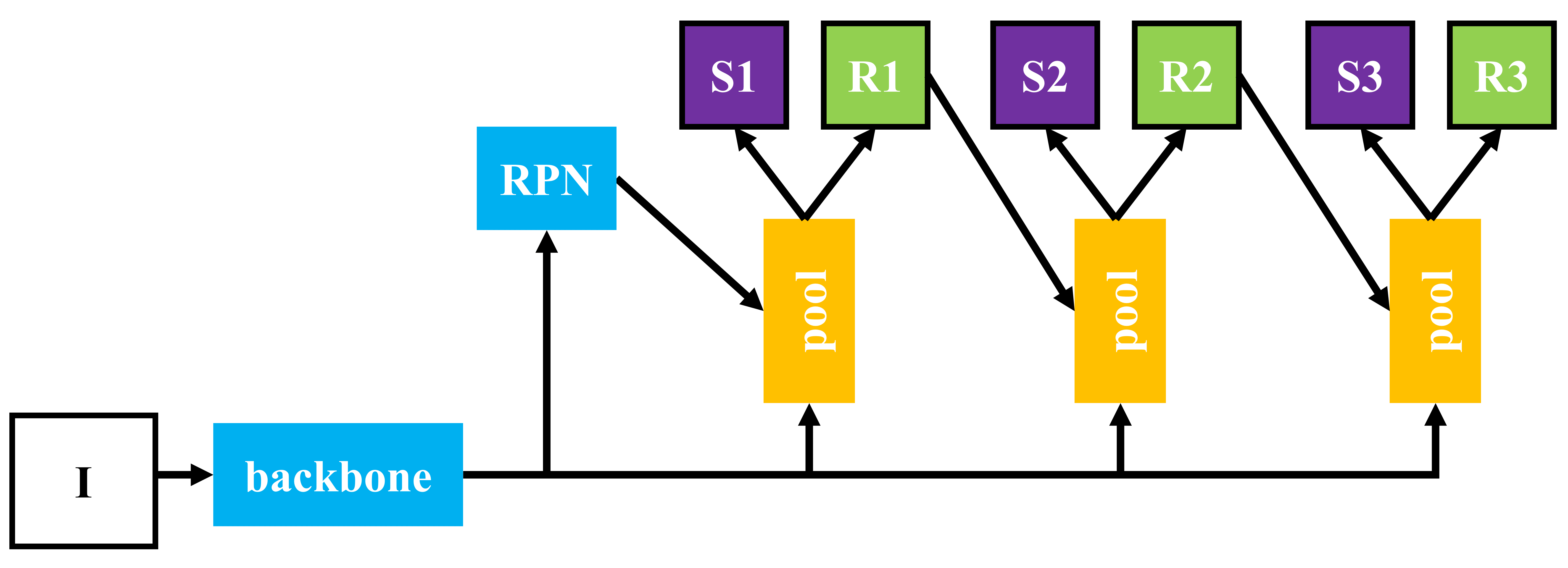}}
\subfigure[Semantic branch]{
\label{fig:figure1b}
\includegraphics[width=5cm]{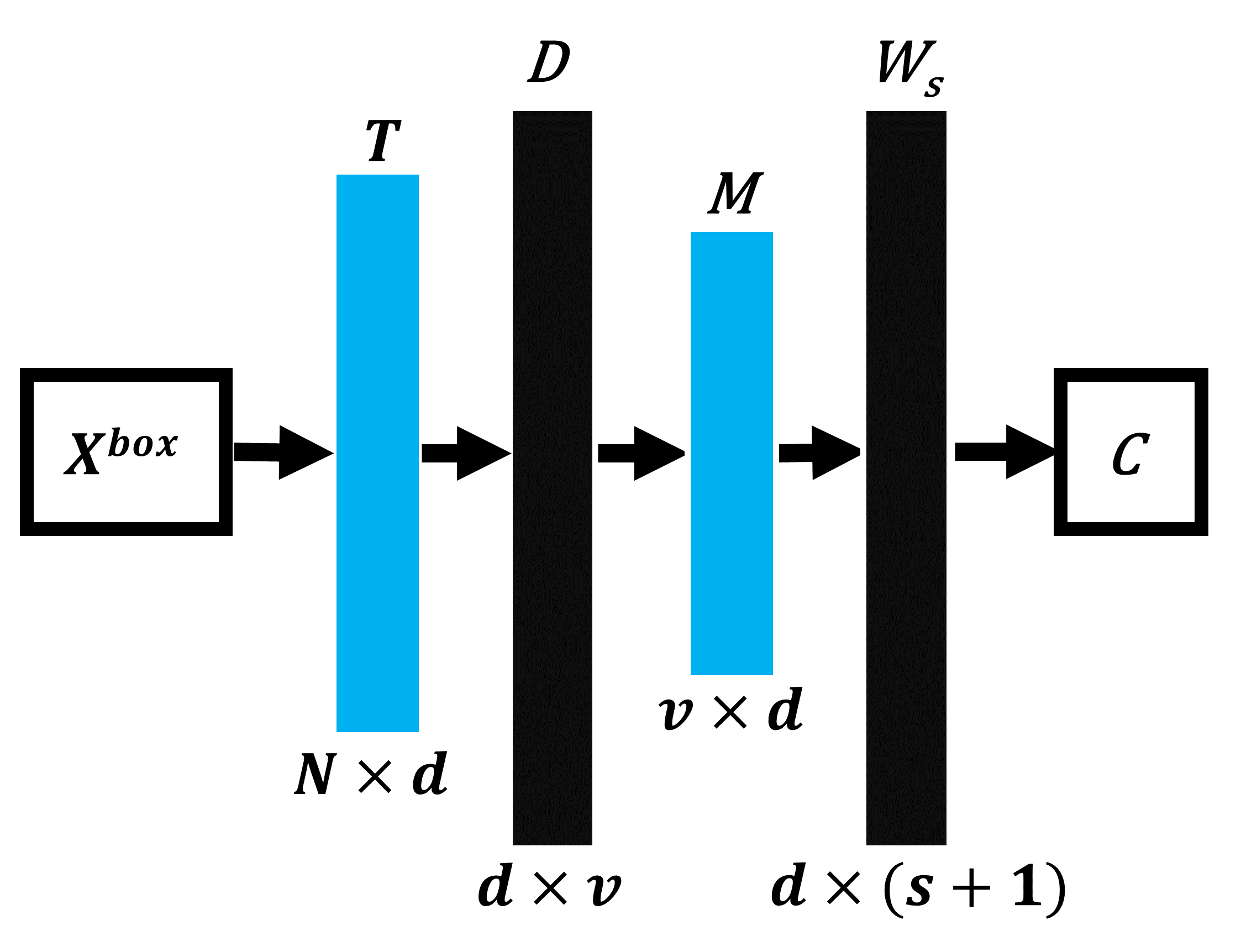}}
\caption{The architecture for Cascade Semantic R-CNN. (a) is the overview architecture and (b) indicates the details for semantic branch. In figure (b), {\it $T$}, {\it $M$} are trainable FC layers and {\it $D$}, {\it $W_s$} are fixed FC layers. For an input image {\it I}, a backbone network (ResNet) is used to obtain the features. Then these features will be forwarded to the Region Proposal Network (RPN) to generate a set of object proposals. After we use a RoI pooling layer to map the proposals' features to a set of fix size objective features, we forward them through the semantic branches (purple {\it {S1}},{\it {S2}} and {\it {S3}}) and the regression branch (green {\it {R1}},{\it {R2}} and {\it {R3}}) in 3 cascade stages to get category scores and bounding boxes for objects.}
\label{fig:figure1}
\end{figure}
%------------------------------------------------------------------------- 
\section{Background Learnable Cascade}
% 开头先总结下BLC方案，说明将要分成以下几个点展开：
% 1. zero-shot Cascade 
% 1.1 semantic branch
% 1.2 baseline zero-shot cascade detection
% 1.3 semantic information flow
% 2. BLRPN
In this section, we elaborate Background Learnable Cascade (BLC). We first introduce our semantic branch about learning the alignment between the visual and semantic information. Then we introduce Cascade Semantic R-CNN which integrates our semantic branch with a multi-stage cascade structure. Since Cascade Semantic R-CNN does not use the semantic information between each stage, we develop semantic information flow structure via incorporating a direct path to reinforce the information flow among semantic branches. Moreover, in consideration of further reducing the confusion between background and unseen classes, we develop BLRPN to learn a discriminative word-embedding representation for background objects. Finally, we describe the details of training process, loss function and inference settings.
%------------------------------------------------------------------------- 
\subsection{Model Architecture}
%------------------------------------------------------------------------- 
\subsubsection{Semantic Branch.\label{Semantic Branch}}
We propose semantic branch to learn the alignment between the visual and semantic information. The details about our semantic branch denoted as $S$ are illustrated in Fig.~\ref{fig:figure1b}. The basic idea is derived from~\cite{rahman2020improved} which uses the relationship between the visual features and the semantic embedding as the bridge to detect unseen objects. There are four main components in semantic branch. $W_{s} \in \mathbb{R}^{d \times (s+1)}$ is a fixed FC layer, whose parameters are the stacked semantic word vectors of background and seen classes. More specifically, $d$ is the dimension of word vector for each class, $s$ denotes the number of seen classes and $1$ denotes the background class. As shown in Fig.~\ref{fig:figure4}, each class has a corresponding word vector $v_c$ ($1 \times d$ dimension) in $W_{s}$. For background class, we use the mean word vector $v_b = \frac{1}{s}\sum_{c=1}^{s}{v_c}$ in our baseline and this $v_b$ will be improved in our BLRPN. Since the word vector quantity for $W_{s}$ is limited and causes the serious sparsity of semantic representation, we add an external vocabulary $D \in \mathbb{R}^{d \times v}$ to enhance the richness of semantic information, where $v$ is the number of words in this external vocabulary. $D$ is also implemented by a fixed FC layer like $W_s$. To overcome the limitation of fixed semantic representation of $W_{s}$ and $D$, we make an updatable representation by introducing an adjustable FC layer $M$ to semantic branch which can be regarded as an attention mechanism in visual-semantic alignment. With this adaptive $M$ whose dimension is $v \times d$, semantic branch can update the semantic word embedding space to learn a more flexible and reliable alignment. $T \in \mathbb{R}^{N \times d}$ is an FC layer which is used to adjust the dimension of input objective feature $\mathbf{x}^{box}$ to fit the subsequent model. In detail, it transforms $\mathbf{x}^{box}$ from $N$ dimension to $d$ dimension. With these above components, our semantic branch projects the input visual feature tensors to the semantic space and then gets the category score $\mathbf{c}$. The calculation process is summarized as follows: 
\begin{equation}
\begin{split}
    S &= \delta(W_{s}MDT), \\
    \mathbf{c} &= \sigma(S(\mathbf{x}^{box})), \\
               &= \sigma(\delta(W_{s}MDT)\mathbf{x}^{box}). 
\end{split}
\label{con:semantic_branch}
\end{equation}
Where, $\delta(\cdot)$ denotes a $\tanh$ activation function, $\sigma(\cdot)$ is the softmax activation function and $\mathbf{c}$ represents the category score.
%------------------------------------------------------------------------- 
\subsubsection{Cascade Semantic R-CNN.\label{Cascade Semantic R-CNN}}
In order to gradually refine the visual-semantic alignment, we integrate above semantic branch into Cascade R-CNN to develop Cascade Semantic R-CNN. We replace the classification branch of each stage for Cascade R-CNN with our semantic branch, as shown in Fig.~\ref{fig:figure1}. In particular, the semantic branches for each stage do not share parameter weights. This framework progressively refines predictions through the semantic branches and bounding box regression branches. The whole pipeline is summarized as follows: 
\begin{equation}
\begin{split}
    \mathbf{x}_{t}^{box} &= \mathcal{P}(\mathbf{x}, \mathbf{r}_{t-1}),  \qquad  \mathbf{r}_{t} = \mathcal{R}_{t}(\mathbf{x}_{t}^{box}), \\
    \mathbf{c}_{t} &= \sigma(\mathbf{S}_{t}(\mathbf{x}_{t}^{box})) = \sigma(\delta(W_{s}M_{t}DT_{t})\mathbf{x}_{t}^{box}).
\end{split}
\label{con:cas-sem}
\end{equation}
Here, $\mathbf{x}$ represents the visual feature from backbone network which is based on ResNet-50~\cite{he2016deep} and the Feature Pyramid Networks (FPN)~\cite{lin2017feature}. 
$\mathbf{r}_{t-1}$ is the RoIs for $(t-1)$-th stage and $\mathbf{x}^{box}_{t}$ represents the objective feature derived from ${\mathbf{x}}$ and the input RoIs $\mathbf{r}_{t-1}$. $\mathcal{P}(\cdot)$ is a pooling operator and we use RoI Align~\cite{maskrcnn} here. $\mathcal{R}_{t}$ and $\mathcal{S}_{t}$ indicate the bounding box regression branch and the semantic branch at the $t$-th stage, respectively. $\mathbf{c}_{t}$ represents category score predictions for $t$-th stage. This process will be iterated in each stage.
%------------------------------------------------------------------------- 
\begin{figure}[tbp]
\centering
\subfigure[]{
\label{fig:figure2a}
\includegraphics[width=5.5cm]{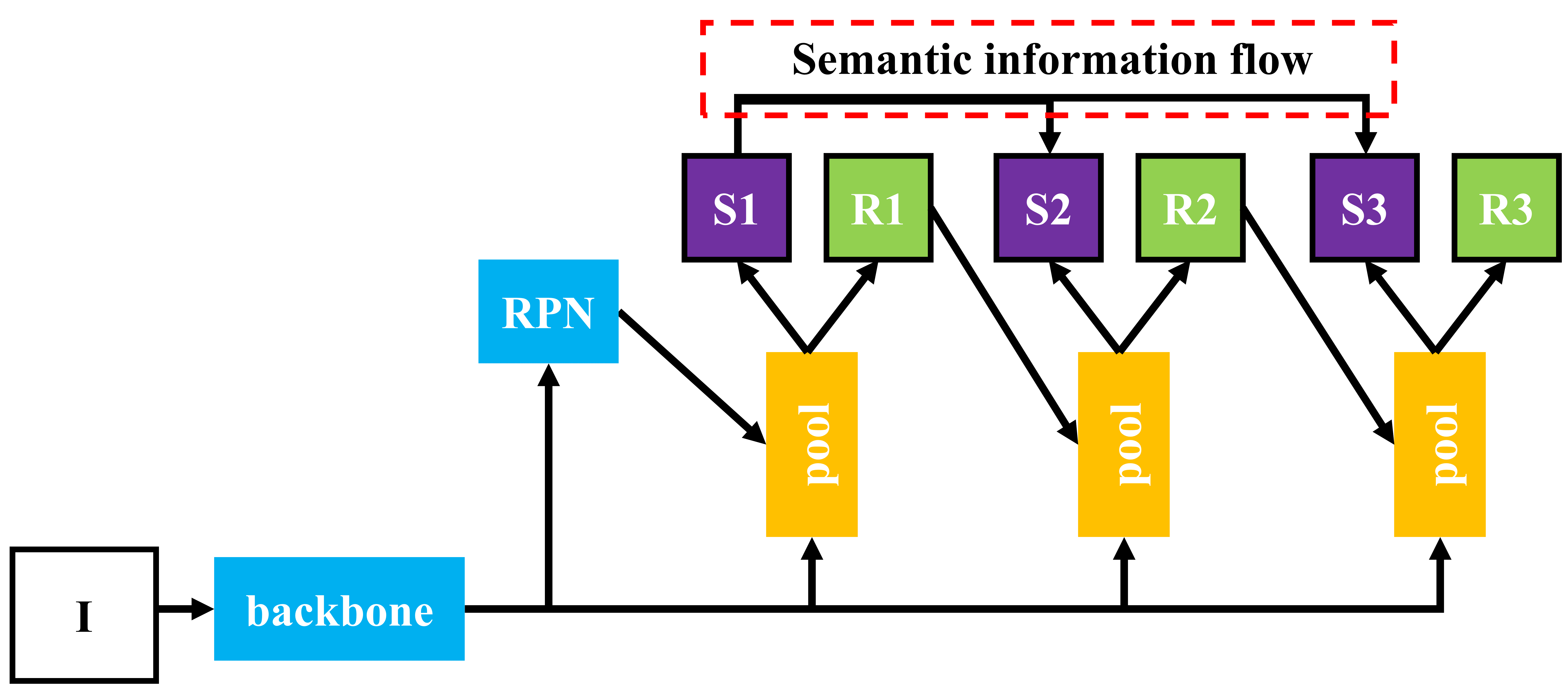}}
\subfigure[]{
\label{fig:figure2b}
\includegraphics[width=5.5cm]{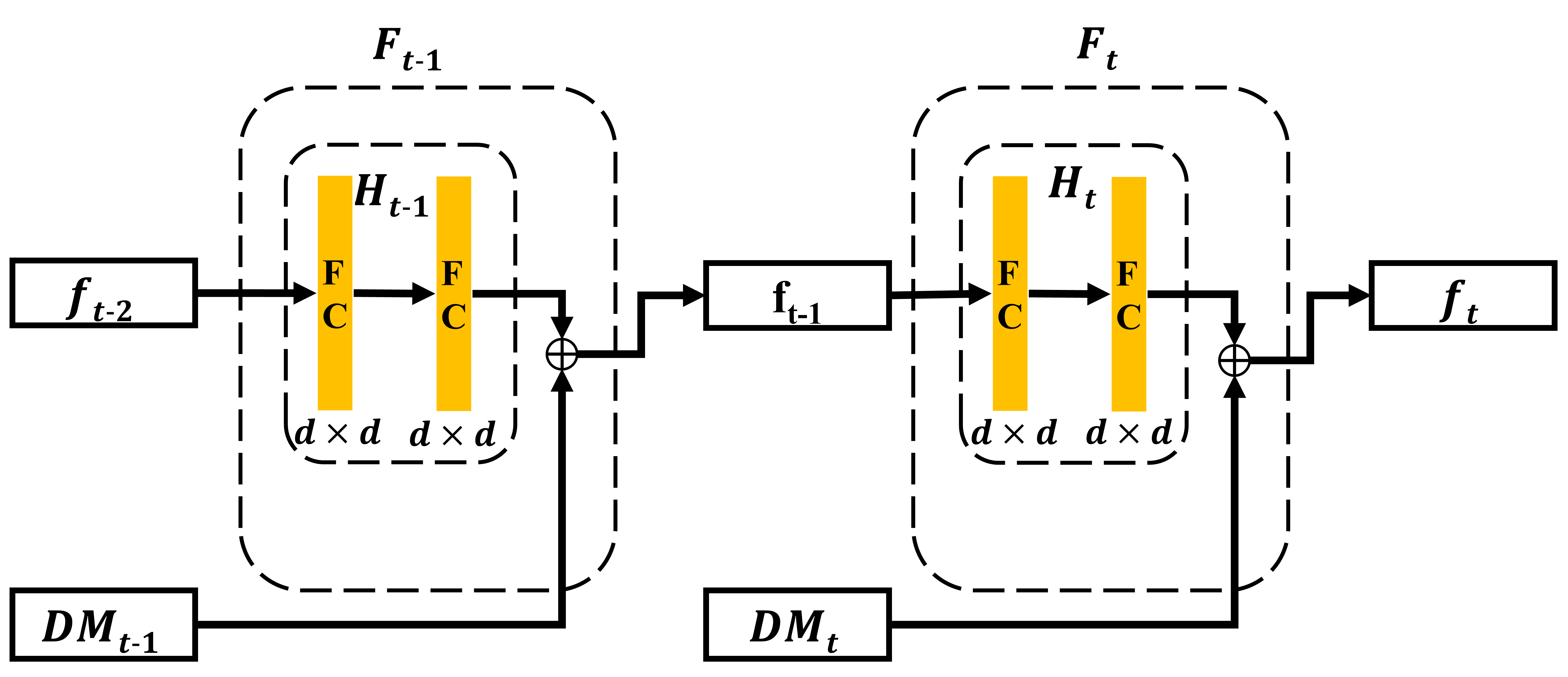}}
\caption{The architecture for semantic information flow. (a) indicates adding semantic information flow into Cascade Semantic R-CNN and (b) shows the details of semantic information flow.}
\label{fig:figure2}
\end{figure}
%------------------------------------------------------------------------- 
\subsubsection{Semantic Information Flow.}
In Cascade Semantic R-CNN, the visual-semantic alignment in semantic branches of each stage is purely based on the visual objective features $\mathbf{x}_{t}^{box}$. This design does not have direct information flow between semantic branches for each stage, failing to make full use of the relevance of semantic information in different stages and progressively refine semantic representing. With the aim of making up this issue, we develop a semantic information flow structure between semantic branches among each cascade stage by forwarding the modulated semantic information from previous stages to current stage, as illustrated in Fig.~\ref{fig:figure2}. We show the calculation process for semantic information flow as follows:
\begin{equation}
\begin{split}
    \mathbf{f}_{1} &= DM_{1} \\
    \mathbf{f}_{2} &= \mathcal{F}_{2}(\mathbf{f}_{1}, DM_{2}) \\
    &\vdots \\
    \mathbf{f}_{t} &= \mathcal{F}_{t}(\mathbf{f}_{t-1}, DM_{t})).
\end{split}
\end{equation}
 Where, $\mathbf{f}_{t}$ represents the semantic information for $t$-th stage derived from $\mathcal{F}_{t}$ which combines the semantic information of current stage and the preceding one. $DM_{t}$ indicates the local semantic information for $t$-th stage.  $\mathcal{F}$ is a function which fuses the semantic information for last stage and current stage with two steps. First, modulating the input semantic information for preceding stage $\mathbf{f}_{t-1}$ with two FC layers $\mathcal{H}_{t}$. Then, adding this modulated feature with the semantic information of current stage $DM_{t}$ in an element-wise manner. The calculation details for $\mathcal{F}$ in $t$-th stage are:
\begin{equation}
\begin{split}
    \mathcal{F}_{t}(\mathbf{f}_{t-1}, DM_{t})) = \mathcal{H}_{t}(\mathbf{f}_{t-1}) + DM_{t}
\end{split}
\end{equation}
After adding the semantic information flow into Cascade Semantic R-CNN, the calculation process for $\mathbf{c}_{t}$ in Equation~\ref{con:cas-sem} will be changed with replacing original $DM_{t}$ with new $\mathbf{f}_{t}$:
\begin{equation}
\begin{split}
   \mathbf{c}_{t} &= \sigma(\mathbf{S}_{t}(\mathbf{x}_{t}^{box})) = \sigma(\delta(W_{s}\mathbf{f}_{t}T_{t})\mathbf{x}_{t}^{box}).
\end{split}
\label{con:eq6}
\end{equation}
The semantic features will benefit from this approach and can help to learn a robust visual-semantic alignment and improve zero shot detection performance.
%------------------------------------------------------------------------- 
\begin{figure}
\centering
\subfigure[BLRPN]{
\label{fig:figure3a}
\includegraphics[width=6cm]{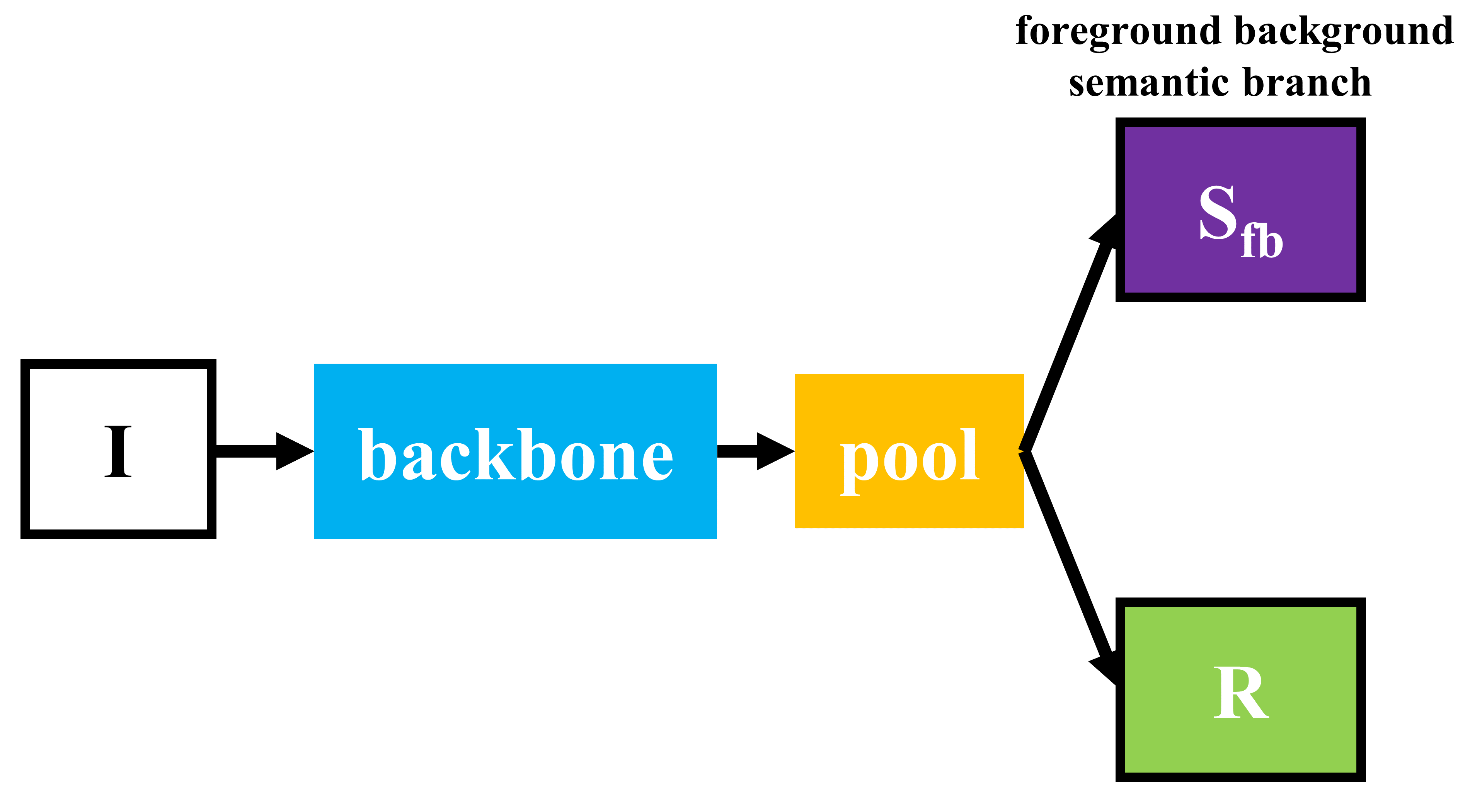}}
\subfigure[foreground-background semantic branch]{
\label{fig:figure3b}
\includegraphics[width=5.5cm]{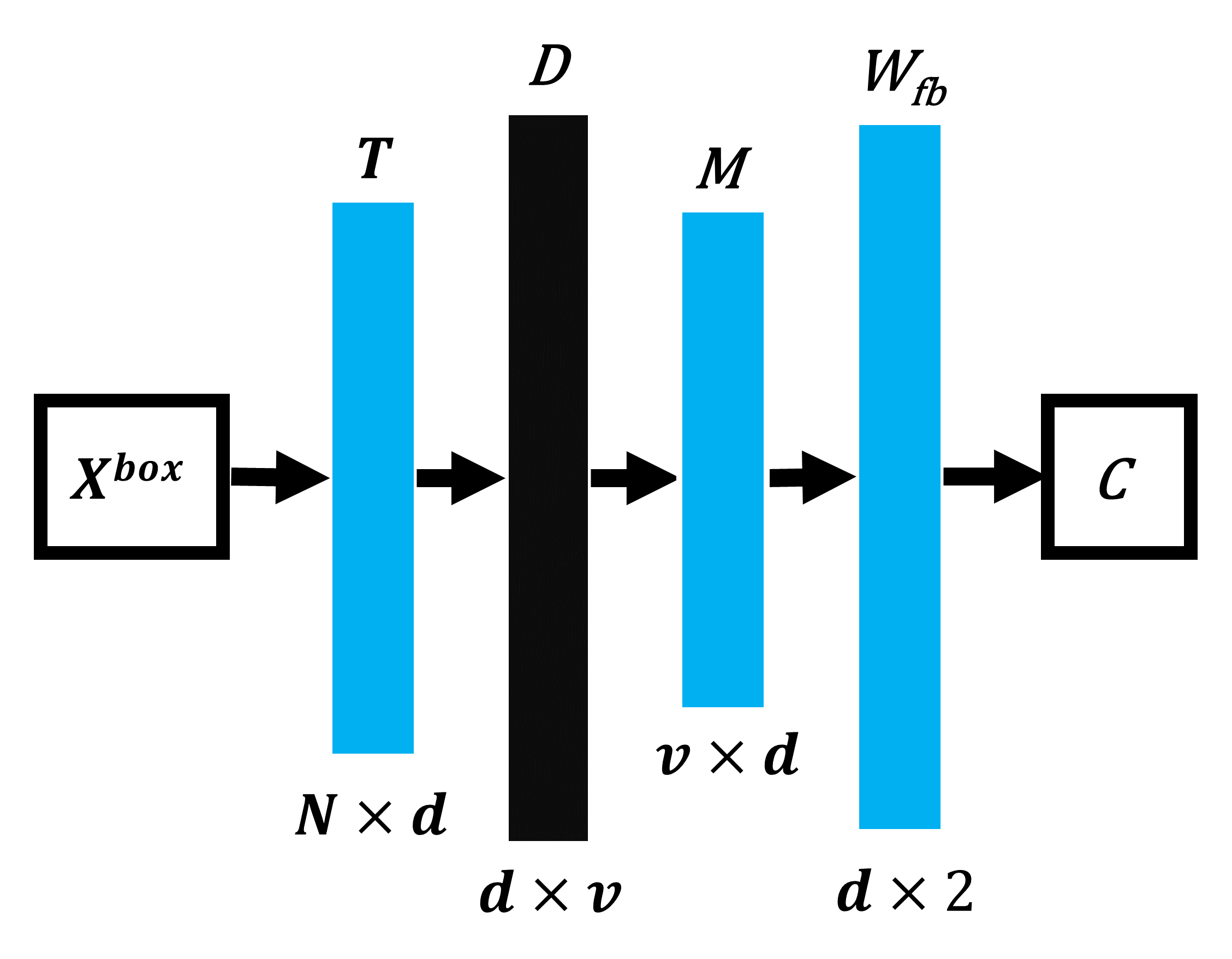}}
\caption{The architecture for BLRPN. (a) is the overview architecture and (b) indicates the details about foreground-background semantic branch $\it {S}_{fb}$. In $\it {S}_{fb}$, $\it {D}$ is fixed while $\it {T}$, $\it {M}$ and $\it {W_{fb}}$ are trainable FC layers. $\it c$ is the foreground background binary classification score.}
\label{fig:figure3}
\end{figure}
%------------------------------------------------------------------------- 
\subsubsection{Background Learnable RPN (BLRPN). \label{BLRPN}}
\begin{figure}[tbp]
\centering
\includegraphics[width=12cm]{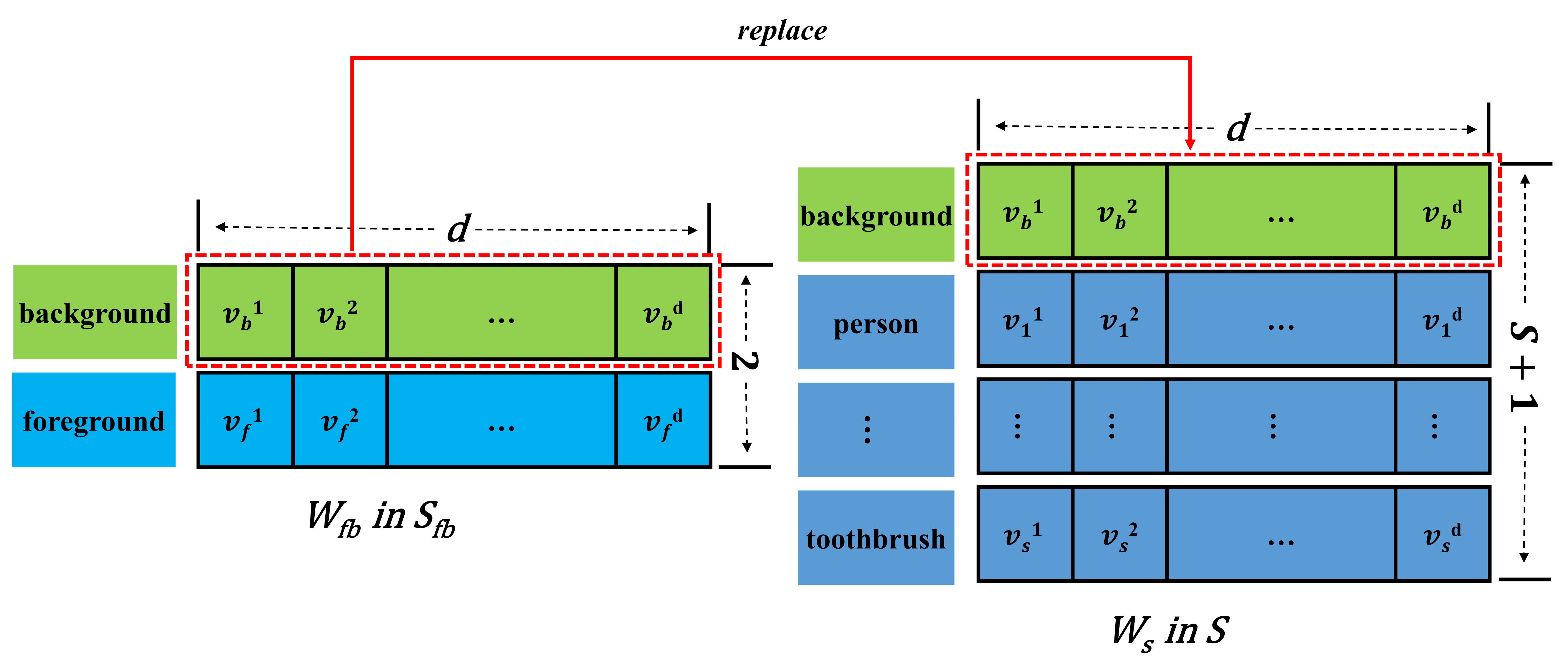}
\caption{$W_s$ is the word vectors for background and other seen classes, it includes 1 background class and s seen classes, each class has a $1 \times d$  dimensional word vector. $W_{fb}$ is the word vectors for background and foreground classes, it includes 1 background class and 1 foreground class, each class has a $1 \times d$ word vector. Here, we replace the $v_b$ in $W_s$ with that in $W_{fb}$ learned from BLRPN.}
\label{fig:figure4}
\end{figure}

In Cascade Semantic R-CNN, the $W_{s}$ in semantic branch adopts a coarse mean word vector $v_b$ for background class, which may not reasonably represent the background class and further reduce the confusion between background and unseen classes. We need a new background semantic vector to replace the old one because this ``replace" strategy can avoid modifying Cascade Semantic R-CNN structure and introducing extra computation. Since the background visual concept is very complex, the better idea is to learn background semantic vector from various background visual data. In order to ensure that the learned background class word vector can directly replace the original one, the learning process needs to be consistent with the process it Cascade Semantic R-CNN.
Based on above analysis, we develop Background Learnable RPN to learn this new background semantic vector and use it to replace the coarse one in $W_{s}$. In Fig.~\ref{fig:figure3}, we develop a foreground-background semantic branch ${S}_{fb}$ and integrate it into the original RPN. ${S}_{fb}$ is modified from our semantic branch for consistency, and the details are illustrated in Fig.~\ref{fig:figure3b}. The only difference between ${S}_{fb}$ and semantic branch ${S}$ is that the $W_s$ in ${S}$ is replaced by the $W_{fb}$ in ${S}_{fb}$. We implement $W_{fb}$ with an FC layer without bias and make it trainable. The parameters of $W_{fb} \in \mathbb{R}^{d \times 2}$ contain two word vectors, one is $v_b$ for background class and the other is $v_f$ for foreground class, so $v_b$ as the new background word vector will be updated during training. $v_f$ is initialized with a uniform random distribution and the $v_b$ is initialized with the mean word vectors for all seen classes, which is the same as $W_{s}$. During training, we feed the visual features derived from the backbone network to the foreground-background branch and get the foreground-background classification score. The details are:
\begin{equation}
\begin{split}
    S_{fb} &= \delta(W_{fb}MDT), \\
    \mathbf{c} &= \sigma(S_{fb}(\mathbf{x}^{box})), \\
               &= \sigma(\delta(W_{fb}MDT)\mathbf{x}^{box}). 
\end{split}
\label{con:blrpn}
\end{equation}
After calculating the loss, we back propagate all gradients to update trainable parameters includes $W_{fb}$. $W_{fb}$ will be updated means that we can learn the target background class semantic vector $v_b$ in the course of training BLRPN. As shown in Fig.~\ref{fig:figure4}, we use this new $v_b$ to replace the old one for background class in $W_{s}$. Finally, we retrain our Cascade Semantic R-CNN model with this new $W_{s}$ and effectively improve the performance for unseen objects. Overall, BLRPN learns the new $v_b$ by establishing the alignment between visual concepts and semantic representation of background classes.
%------------------------------------------------------------------------- 
%------------------------------------------------------------------------- 
\subsection{Learning}
%------------------------------------------------------------------------- 
\subsubsection{Training Process.}
% 这里要介绍训练两个模型的过程
Compared with previous achievements~\cite{bansal2018zero,demirel2018zero,rahman2018zero,zhu2019zero} needing multi-step training and pre-trained weights on seen or unseen data, the training process of our model is very simple and convenient with a two step manner. First we train BLRPN to get $v_b$ and use it to obtain a new $W_s$. Then we train our Cascade Semantic R-CNN equipped with semantic information flow with this new $W_s$. It needs to be emphasized that we only adopt the ImageNet pre-trained weights in the above training processes without any pre-trained weights of seen-class data.
%------------------------------------------------------------------------- 
\subsubsection{Loss Function.}
First, we introduce the loss function of Cascade Semantic R-CNN. In each stage $t$ for Cascade Semantic R-CNN, the box regression branch predicts the RoIs $\mathbf{r}_{t}$ and the semantic branch predicts category score $c_{t}$. The loss function $L_{cs}$ is:
\begin{equation}
\begin{split}
    L_{cs} = \sum_{t=1}^{3}\alpha_{t}(L_{t}^{reg}+L_{t}^{sem}), \\
    L_{t}^{reg}(\mathbf{r}_{t}, \widehat{\mathbf{r}}_{t}) = \ell_{1}(\mathbf{r}_{t}, \widehat{\mathbf{r}}_{t}), \\ 
    L_{t}^{sem}(c_{t}, \widehat{c}_{t}) = CE(c_{t}, \widehat{c}_{t}).
\end{split}
\end{equation}
Here, $L_{t}^{sem}$ represents classification loss for semantic branch which adopts cross-entropy (CE) loss function. $L_{t}^{reg}$ is the loss of the boxes predictions at stage t, which uses smooth $\ell{1}$ loss. The coefficient $\alpha_{t}$ is the loss weight for each stage, we follow the settings in Cascade R-CNN~\cite{cascadercnn} and set $\alpha_{t}$ to [1,0.5,0.25] for 3 stages. 

The loss function of BLRPN denoted as $L_{blrpn}$ is consists of the classification loss $L^{fbsem}$ in foreground-background semantic branch and the box regression loss $L^{reg}$ in regression branch:
\begin{equation}
\begin{split}
    L_{blrpn} = L^{reg} + L^{fbsem}, \\
    L^{reg}(\mathbf{r}, \widehat{\mathbf{r}}) = \ell_{1}(\mathbf{r}, \widehat{\mathbf{r}}), \\ 
    L^{fbsem}(c, \widehat{c}) = CE(c, \widehat{c}).
\end{split}
\end{equation}
%------------------------------------------------------------------------- 
\subsubsection{Inference.}
We forward the input images through Cascade Semantic R-CNN to get the boxes and categories for all objects, then we apply Non-Maximum Suppression (NMS) to get the final results. In addition to the original inference process for the seen class in Equation~\ref{con:eq6}, we add an extra calculation process to inference unseen objects. The extra process is as follows:
\begin{align}
\mathbf{c}_{unseen} = 
    W_{u}W_{s}^\mathsf{T}\sigma(\delta(W_{s}fT)\mathbf{x}^{box}).
\end{align}
Where, $W_{u} \in \mathbb{R}^{d \times (u+1)}$ denotes the stacking word vectors for background and unseen classes, $u$ indicates the number of unseen classes. The other components are same as Equation~\ref{con:eq6}. For an input object feature $\mathbf{x}^{box}$, we first map this visual feature to the category probability of seen classes. Then we use the transpose of $W_{s}$ to transform this probability back to semantic space, finally we get unseen category score from the semantic space through $W_{u}$. For GZSD task, we simultaneously execute the above two reasoning process, so as to achieve the simultaneous reasoning of seen and unseen objects.
%------------------------------------------------------------------------- 
\section{Experiments}
\subsection{Datasets}
% \subsection{Preparing Setup}
% \subsubsection{Datasets.}
We perform experiments on MS-COCO dataset~\cite{lin2014microsoft}. MS-COCO (2014) includes 82783 training images and 40504 validation images with 80 classes. We follow the datasets settings in~\cite{bansal2018zero} and~\cite{rahman2020improved} for MS-COCO. We divide the dataset with two different splits: (i) 48 seen classes an 17 unseen classes; (ii) 65 seen classes and 15 unseen classes. The seen classes are training set and unseen classes are test set. Both splits remove all images from the training set which contain any object from seen classes. Specially, the images for unseen classes in test set still have objects for seen classes in order to maintain the number of samples in the test set. Following~\cite{rahman2020improved}, we use extra vocabulary from NUS-WIDE~\cite{chua2009nus} and remove MS-COCO classes names and all tags with no word-vectors. We use a 300 dimensional word2vec~\cite{mikolov2013distributed} with a $\ell_{2}$ normalization for MS-COCO classes and extra vocabulary. 
%------------------------------------------------------------------------- 
\subsection{Evaluation Protocol}
We report the evaluation results on ZSD and GZSD task like previous work~\cite{bansal2018zero,rahman2020improved} over two splits for MS-COCO. We use recall and mAP as metrics, these metrics for boxes are all evaluated across IoU thresholds in 0.4, 0.5 and 0.6. In particular, the evaluation for recall is based on Recall@K~\cite{bansal2018zero}, which means the recall when only the top K detections are selected from an image, we set K to 100 by following the settings in~\cite{bansal2018zero}.
%------------------------------------------------------------------------- 
\subsection{Implementation Details}
In all experiments, we adopt ResNet-50~\cite{he2016deep} as the backbone network with FPN~\cite{lin2017feature}. We train all models with 4 GPUs (two images per GPU) for 12 epochs with a SGD optimizer which momentum is 0.9 and weight-decay is 0.0001. The initial learning rate for the optimizer is set to 0.01, and decreased by 0.1 after 8 and 11 epochs. The long edge and short edge of images are resized to 1333 and 800 without changing the aspect ratio. We use horizontal flip during training and the multi-scale for training is set to [400, 1400]. We implement our model in PyTorch~\cite{paszke2017automatic} and the pre-trained model is from PyTorch official model zoo.
%------------------------------------------------------------------------- 
\setlength{\tabcolsep}{6pt}
\begin{table}[tbp]
\begin{center}
\caption{Comparison of the proposed BLC with the previous state-of-the-art zsd work on two splits of COCO. Seen/Unseen refers to the split of datasets. The proposed BLC can achieve 10.6 mAP and 48.87 Recall@100 for 48/17 split, 14.7 mAP and 54.68 Recall@100 for 65/15 split, significantly surpasses all other work. “ms” indicates multi-scale training and test.}
\label{table:benchmark result}
\begin{tabular}{cccccc}
% \noalign{\smallskip}
\toprule
% \noalign{\smallskip}
% \toprule
\multirow{2}{*}{Method} & \multirow{2}{*}{Seen/Unseen} & \multicolumn{3}{c}{Recall@100} & mAP \\
\cmidrule(lr){3-5} \cmidrule(lr){6-6}
~ & ~ & 0.4 & 0.5 & 0.6 & 0.5 \\
\midrule
SB~\cite{bansal2018zero} & 48/17 & 34.46 & 22.14 & 11.31 & 0.32 \\
DSES~\cite{bansal2018zero} & 48/17 & 40.23 & 27.19 & 13.63 & 0.54 \\
TD~\cite{li2019zero} & 48/17 & 45.50 & 34.30 & 18.10 & - \\
PL~\cite{rahman2020improved} & 48/17 & - & 43.59 & - & 10.10 \\
\hline
\textbf{BLC } & 48/17 &  \textbf{49.63} &  \textbf{46.39} &  \textbf{41.86} & 9.90 \\
\textbf{BLC (ms)} & 48/17 &  \textbf{51.33} & \textbf{48.87} & \textbf{45.03} & \textbf{10.60} \\
\bottomrule
PL~\cite{rahman2020improved} & 65/15 & - & 37.72 & - & 12.40  \\
\hline
\textbf{BLC} & 65/15 & \textbf{54.18} & \textbf{51.65} & \textbf{47.86} & \textbf{13.10} \\
\textbf{BLC (ms)} & 65/15 &  \textbf{57.23} & \textbf{54.68} & \textbf{51.22} & \textbf{14.70} \\
\bottomrule
\end{tabular}
\end{center}
\end{table}
\setlength{\tabcolsep}{1.4pt}
%------------------------------------------------------------------------- 
\subsection{Quantitative Results}
%------------------------------------------------------------------------- 
\subsubsection{Results in Benchmarks.}
We compare Background Learnable Cascade with the state-of-the-art zero-shot detection approaches on two splits of MS-COCO in Table~\ref{table:benchmark result}. We can observe that: (i) for 48/17 split, we compare our approaches with SB~\cite{bansal2018zero}, DSES~\cite{bansal2018zero}, TD~\cite{li2019zero} and PL~\cite{rahman2020improved}. Our BLC surpasses all of them in Recall@100 and mAP, brings up to 33.72\% ($4\times$) and 10.28\% ($33\times$) gain in terms of Recall@100 and mAP; (ii) for 65/15 split, compared with PL~\cite{rahman2020improved}, our BLC brings 16.96\% gain for Recall@100 and 2.3\% improvement for mAP.
Moreover, in other previous works, Recall@100 drops severely as IoU threshold increasing while our BLC can still maintain a high Recall@100 indicating our approach is more robust for stringent IoU threshold.
%------------------------------------------------------------------------- 
\subsubsection{Component-wise Analysis.}
We investigate the contributions of the main components for BLC. ``Cascade Semantic" means the baseline Cascade Semantic R-CNN, ``Semantic Flow" denotes the semantic information flow, ``BLRPN" represents the new background class word vector learned from our background learnable region proposal network. The results for 48/17 and 65/15 splits are shown in Table~\ref{table:component-wise}, respectively. 
% Compared with our baseline: (i) for 48/17 split, semantic information flow improves the Recall@100 by 2.98\%, BLRPN contributes 6.66\% and all components lead to a further improvement by 7.64\% together; (ii) for 65/15 split, semantic information flow improves Recall@100 by 1.77\%, BLRPN brings 3.75\% and leads to 4.37\% together.
%-------------------------------------------------------------------------
\setlength{\tabcolsep}{2pt}
\begin{table}[tbp]
\begin{center}
\caption{Effects of each component in our work. Results are reported on 48/17 split and 65/15 split of MS-COCO, respectively.}
\label{table:component-wise}
% \resizebox{\textwidth}{!}{
\begin{tabular}{c|ccccccc}
\toprule
 \multirow{2}{*}{ } & \multirow{2}{*}{Cascade Semantic} & \multirow{2}{*}{Semantic Info} & \multirow{2}{*}{BLRPN} & \multicolumn{3}{c}{Recall@100} & mAP \\
\cmidrule(lr){5-7} \cmidrule(lr){8-8}
 ~ & ~ & ~ & ~ & 0.4 & 0.5 & 0.6 & 0.5 \\
\cmidrule(lr){1-8}
\multirow{4}{*}{\rotatebox{90}{48/17}} & \checkmark &  &  & 40.96 & 38.75 & 35.25 & 9.3 \\
~ & \checkmark &  \checkmark &  & 43.84 & 41.73 & 38.11 & 9.5 \\
~ & \checkmark &  &  \checkmark & 48.52 & 45.41 & 41.04 & 9.6 \\
~ & \checkmark &  \checkmark &  \checkmark & \textbf{49.63} & \textbf{46.39} & \textbf{41.86} & \textbf{9.9} \\
\bottomrule
\noalign{\smallskip}
\multirow{4}{*}{\rotatebox{90}{65/15}} & \checkmark &  &  & 49.75 & 47.28 & 43.87 & 12.4 \\
~ & \checkmark &  \checkmark &  & 51.49 & 49.05 & 45.07 & 12.7 \\
~ & \checkmark &  &  \checkmark & 53.38 & 51.03 & 47.39 & 12.9 \\
~ & \checkmark &  \checkmark &  \checkmark & \textbf{54.18} & \textbf{51.65} & \textbf{47.86} & \textbf{13.1} \\
\bottomrule
\end{tabular}
% }
\end{center}
\end{table}
\setlength{\tabcolsep}{1.4pt}
%-------------------------------------------------------------------------
\subsubsection{Class-wise Performance.}
We report the Recall@100 on two splits of MS-COCO for each unseen classes in Table~\ref{table:class-wise}. Our BLC makes significant improvement on both splits: (i) for the split of 48/17, BLC substantially boosts baseline in the most of classes. For the classes which are hard to detect, BLC achieves $2.1\times$, $1.6\times$, $3.7\times$, $1.4\times$, $2.1\times$, $1.5\times$ and $2.5\times$ improvement on Recall@100 for ``skateboard", ``cup", ``knife", ``cake", ``keyboard", ``sink" and ``scissors" classes, respectively; (ii) for the split of 65/15, BLC also obtains further improvement compared with baseline. We also note that BLC is unable to detect any true positive for the class ``umbrella" and ``tie", the Recall@100 rate for the class ``hair drier" is also unsatisfying. The main reason is that there are fewer classes are semantically similar with these poor classes in training dataset, which makes them difficult to detect.
%-------------------------------------------------------------------------
\setlength{\tabcolsep}{4pt}
\begin{table}[tbp]
\begin{center}
\caption{Class-wise Recall@100 for 48/17 and 65/15 splits of MS-COCO with the IoU threshold is 0.5. Our BLC achieves significant improvement in most of unseen classes compared
with Cascade Semantic R-CNN baseline.}
\label{table:class-wise}
\begin{tabular}{c}
     48/17 split of MS-COCO
\end{tabular}
\resizebox{\linewidth}{!}{
\begin{tabular}{|c|c|c|c|c|c|c|c|c|c|c|c|c|c|c|c|c|c|c|}
\hline
Method & \rotatebox{90}{Overall}& \rotatebox{90}{bus} & \rotatebox{90}{dog} & \rotatebox{90}{cow} & \rotatebox{90}{elephant} & \rotatebox{90}{umbrella} & \rotatebox{90}{tie} & \rotatebox{90}{skateboard} & \rotatebox{90}{cup} & \rotatebox{90}{knife} & \rotatebox{90}{cake} & \rotatebox{90}{couch} & \rotatebox{90}{keyboard} & \rotatebox{90}{sink} & \rotatebox{90}{scissors} & \rotatebox{90}{airplane} & \rotatebox{90}{cat} & \rotatebox{90}{snowboard} \\
\hline
baseline & 38.73 & 72.9 & \textbf{94.6} & 67.3 & 68.1 & 0.0 & 0.0 & 19.9 & 24.0 & 12.4 & 24.0 & 63.7 & 11.6 & 9.2 & 8.3 & 48.3 & 70.7 & \textbf{63.4} \\
\hline
BLC & \textbf{46.39} & \textbf{77.4} & 88.4 & \textbf{71.9} & \textbf{77.2} & 0.0 & 0.0 & \textbf{41.7} & \textbf{38.0} & \textbf{45.6} & \textbf{34.3} & \textbf{65.2} & \textbf{23.8} & \textbf{14.1} & \textbf{20.8} & 48.3 & \textbf{79.9} & 61.8 \\
\hline
\end{tabular}
}
\begin{tabular}{c}
     ~ \\
     65/15 split of MS-COCO
\end{tabular}
\resizebox{\linewidth}{!}{
\begin{tabular}{|c|c|c|c|c|c|c|c|c|c|c|c|c|c|c|c|c|}
% \noalign{\smallskip}
% \toprule
% \noalign{\smallskip}
\hline
Method & \rotatebox{90}{Overall}& \rotatebox{90}{airplane} & \rotatebox{90}{train} & \rotatebox{90}{parking meter} & \rotatebox{90}{cat} & \rotatebox{90}{bear} & \rotatebox{90}{suitcase} & \rotatebox{90}{frisbee} & \rotatebox{90}{snowboard} & \rotatebox{90}{fork} & \rotatebox{90}{sandwich} & \rotatebox{90}{hot dog} & \rotatebox{90}{toilet} & \rotatebox{90}{mouse} & \rotatebox{90}{toaster} & \rotatebox{90}{hair drier} \\
\hline
baseline & 47.28 & 53.9 & 70.6 & 5.9 & 90.2 & 85.1 & 40.7 & 25.9 & 59.9 & 33.7 & 76.9 & 64.4 & 33.2 & 3.3 & \textbf{64.1} & 1.4 \\
\hline
BLC & \textbf{51.28} & \textbf{58.7} & \textbf{72.0} & \textbf{10.2} & \textbf{96.1} & \textbf{91.6} & \textbf{46.9} & \textbf{44.1} & \textbf{65.4} & \textbf{37.9} & \textbf{82.5} & \textbf{73.6} & \textbf{43.8} & \textbf{7.9} & 35.9 & \textbf{2.7}\\
\hline
\end{tabular}
}
\end{center}
\end{table}
\setlength{\tabcolsep}{1.4pt}
%------------------------------------------------------------------------- 
\setlength{\tabcolsep}{6pt}
\begin{table}[htb]
\begin{center}
\caption{This table shows Recall@100 and mAP (IoU threshold=0.5) for our BLC and other stat of the art over GZSD task. HM denotes the harmonic average for seen and unseen classes.}
\label{table:gzsd result}
\begin{tabular}{cccccccc}
% \noalign{\smallskip}
\toprule
% \noalign{\smallskip}
% \toprule
\multirow{2}{*}{Method} & \multirow{2}{*}{Seen/Unseen} & \multicolumn{2}{c}{seen} & \multicolumn{2}{c}{unseen} & \multicolumn{2}{c}{HM} \\
\cmidrule(lr){3-4} \cmidrule(lr){5-6} \cmidrule(lr){7-8}
~ & ~ & mAP & Recall & mAP & Recall & mAP & Recall \\
\midrule
DSES~\cite{bansal2018zero} & 48/17 & - & 15.02 & - & 15.32 & - & 15.17 \\
PL~\cite{rahman2020improved} & 48/17 & 35.92 & 38.24 & 4.12 & 26.32 & 7.39 & 31.18 \\
\hline
BLC & 48/17 & \textbf{42.10} & \textbf{57.56} &  \textbf{4.50} & \textbf{46.39} & \textbf{8.20} & \textbf{51.37} \\
\bottomrule
PL~\cite{rahman2020improved} & 65/15 & 34.07 & 36.38 & 12.40 & 37.16 & 18.18 & 36.76 \\
\hline
BLC & 65/15 & \textbf{36.00} & \textbf{56.39} & \textbf{13.10} & \textbf{51.65} & \textbf{19.20} & \textbf{53.92}  \\
\bottomrule
\end{tabular}
\end{center}
\end{table}
\setlength{\tabcolsep}{1.4pt}
%-------------------------------------------------------------------------
\subsubsection{Generalized Zero-Shot Detection (GZSD) Results.}
The generalized zero-shot detection task is more realistic that both seen and unseen classes are presented during evaluation. We report the performance for GZSD in Table~\ref{table:gzsd result} under on both splits over MS-COCO. The score threshold is 0.2 for seen classes and 0.05 for unseen classes, respectively. The IoU threshold for mAP is 0.5. Our BLC exceeds other stat-of-the-art methods in terms of mAP and recall@100.
%-------------------------------------------------------------------------
\subsection{Qualitative Results}
\begin{figure}[tbp]
\centering
\includegraphics[width=\linewidth]{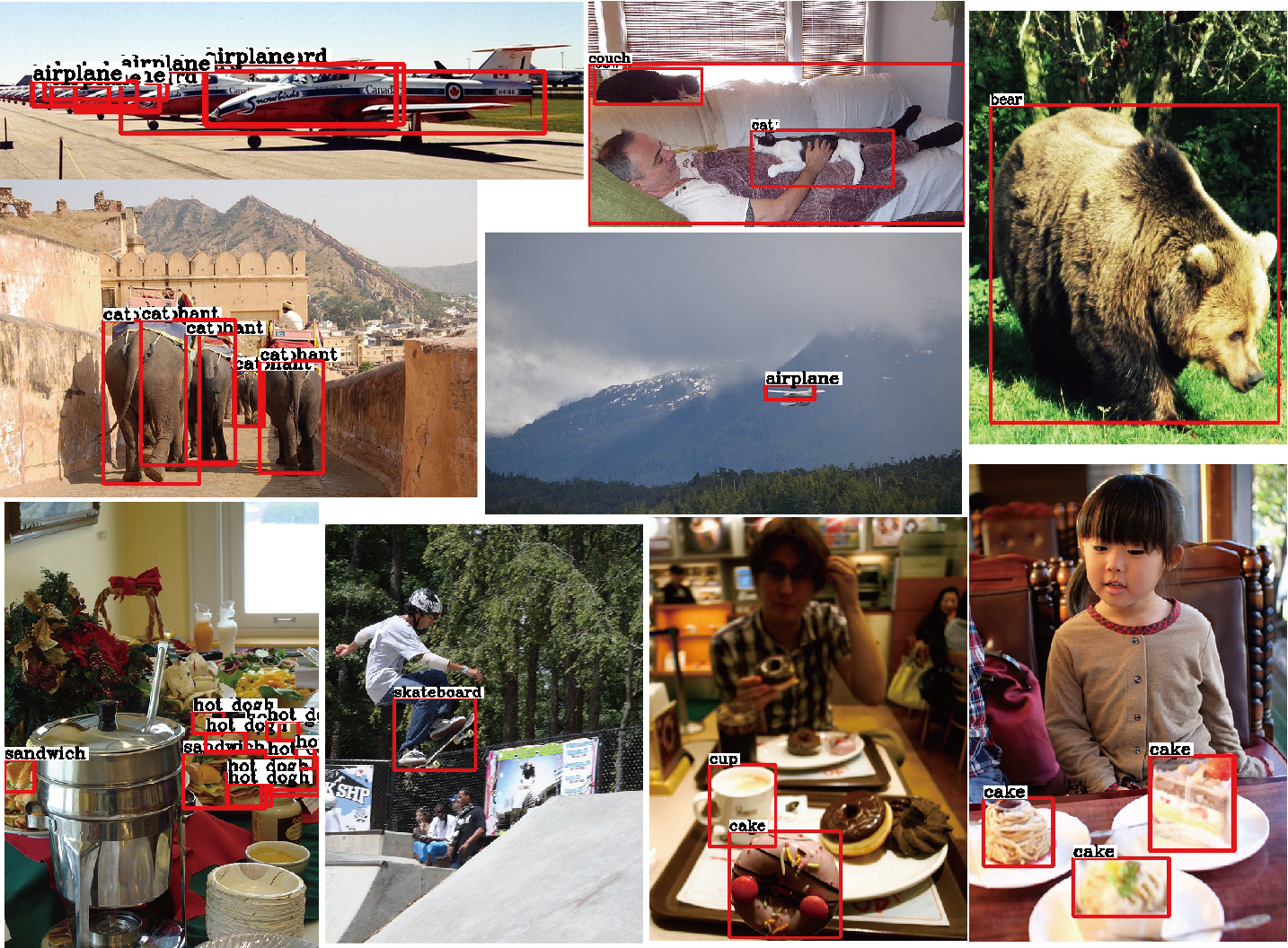}
\caption{Examples for detection results of BLC on 48/17 and 65/15 splits of MS-COCO. All these objects are belong unseen classes.}
\label{fig:figure5}
\end{figure}

For intuitively evaluating the qualitative results, we give some detection results in Fig.~\ref{fig:figure5} for BLC on two splits of MS-COCO. We find that BLC can precisely detect unseen classes under different situations. For example, BLC detects objects under densely packed scenes, e.g., ``airplanes", ``elephants" and ``hot dogs", as well as successfully captures small objects like the tiny ``airplane". It is noteworthy that multiple objects are also detected by BLC from messy background like ``cat" and ``couch". The main issue in BLC is the misclassification for unseen objects which belong to the same meta class due to lacking of enough information to distinguish them, and we can see it as cases of "elephant" and "cat". 
%-------------------------------------------------------------------------
\section{Conclusions}
In this paper, we propose a novel framework for ZSD named Background Learnable Cascade (BLC), which includes Cascade Semantic R-CNN, semantic information flow and BLRPN. Cascade Semantic R-CNN progressively refines the visual-semantic alignment, semantic information flow improves the semantic feature learning and BLRPN learns a appropriate word vector for background class to reduce the confusion between background and unseen classes. Experiments in two splits of MS-COCO show that BLC outperforms several state of the art under both ZSD and GZSD tasks.

\section*{Acknowledgement}
The paper is supported by the National Natural Science Foundation of China (NSFC) under Grant No. 61672498 and the National Key Research and Development Program of China under Grant No. 2016YFC0302300.
%-------------------------------------------------------------------------
%===========================================================
\bibliographystyle{splncs}
\bibliography{egbib}

\begin{thebibliography}{10}

\bibitem{zhang2016zero1}
Zhang, Z., Saligrama, V.:
\newblock Zero-shot learning via joint latent similarity embedding.
\newblock In: proceedings of the IEEE Conference on Computer Vision and Pattern
  Recognition. (2016)  6034--6042

\bibitem{nilsback2008automated}
Nilsback, M.E., Zisserman, A.:
\newblock Automated flower classification over a large number of classes.
\newblock In: 2008 Sixth Indian Conference on Computer Vision, Graphics \&
  Image Processing, IEEE (2008)  722--729

\bibitem{russakovsky2015imagenet}
Russakovsky, O., Deng, J., Su, H., Krause, J., Satheesh, S., Ma, S., Huang, Z.,
  Karpathy, A., Khosla, A., Bernstein, M.,  et~al.:
\newblock Imagenet large scale visual recognition challenge.
\newblock International journal of computer vision \textbf{115} (2015)
  211--252

\bibitem{welinder2010caltech}
Welinder, P., Branson, S., Mita, T., Wah, C., Schroff, F., Belongie, S.,
  Perona, P.:
\newblock Caltech-ucsd birds 200.
\newblock (2010)

\bibitem{bansal2018zero}
Bansal, A., Sikka, K., Sharma, G., Chellappa, R., Divakaran, A.:
\newblock Zero-shot object detection.
\newblock In: Proceedings of the European Conference on Computer Vision (ECCV).
  (2018)  384--400

\bibitem{demirel2018zero}
Demirel, B., Cinbis, R.G., Ikizler-Cinbis, N.:
\newblock Zero-shot object detection by hybrid region embedding.
\newblock arXiv preprint arXiv:1805.06157 (2018)

\bibitem{rahman2018zero}
Rahman, S., Khan, S., Porikli, F.:
\newblock Zero-shot object detection: Learning to simultaneously recognize and
  localize novel concepts.
\newblock In: Asian Conference on Computer Vision, Springer (2018)  547--563

\bibitem{zhu2019zero}
Zhu, P., Wang, H., Saligrama, V.:
\newblock Zero shot detection.
\newblock IEEE Transactions on Circuits and Systems for Video Technology (2019)

\bibitem{rahman2020improved}
Rahman, S., Khan, S., Barnes, N.:
\newblock Improved visual-semantic alignment for zero-shot object detection.
\newblock 34th AAAI Conference on Artificial Intelligence (2020)

\bibitem{li2019zero}
Li, Z., Yao, L., Zhang, X., Wang, X., Kanhere, S., Zhang, H.:
\newblock Zero-shot object detection with textual descriptions.
\newblock In: Proceedings of the AAAI Conference on Artificial Intelligence.
  Volume~33. (2019)  8690--8697

\bibitem{zhao2020gtnet}
Zhao, S., Gao, C., Shao, Y., Li, L., Yu, C., Ji, Z., Sang, N.:
\newblock Gtnet: Generative transfer network for zero-shot object detection.
\newblock arXiv preprint arXiv:2001.06812 (2020)

\bibitem{zhu2020don}
Zhu, P., Wang, H., Saligrama, V.:
\newblock Don't even look once: Synthesizing features for zero-shot detection.
\newblock In: Proceedings of the IEEE/CVF Conference on Computer Vision and
  Pattern Recognition. (2020)  11693--11702

\bibitem{cascadercnn}
Cai, Z., Vasconcelos, N.:
\newblock Cascade r-cnn: Delving into high quality object detection.
\newblock In: Proceedings of the IEEE conference on computer vision and pattern
  recognition. (2018)  6154--6162

\bibitem{bendale2016towards}
Bendale, A., Boult, T.E.:
\newblock Towards open set deep networks.
\newblock In: Proceedings of the IEEE conference on computer vision and pattern
  recognition. (2016)  1563--1572

\bibitem{changpinyo2016synthesized}
Changpinyo, S., Chao, W.L., Gong, B., Sha, F.:
\newblock Synthesized classifiers for zero-shot learning.
\newblock In: Proceedings of the IEEE conference on computer vision and pattern
  recognition. (2016)  5327--5336

\bibitem{elhoseiny2013write}
Elhoseiny, M., Saleh, B., Elgammal, A.:
\newblock Write a classifier: Zero-shot learning using purely textual
  descriptions.
\newblock In: Proceedings of the IEEE International Conference on Computer
  Vision. (2013)  2584--2591

\bibitem{frome2013devise}
Frome, A., Corrado, G.S., Shlens, J., Bengio, S., Dean, J., Ranzato, M.,
  Mikolov, T.:
\newblock Devise: A deep visual-semantic embedding model.
\newblock In: Advances in neural information processing systems. (2013)
  2121--2129

\bibitem{jain2014multi}
Jain, L.P., Scheirer, W.J., Boult, T.E.:
\newblock Multi-class open set recognition using probability of inclusion.
\newblock In: European Conference on Computer Vision, Springer (2014)  393--409

\bibitem{kodirov2017semantic}
Kodirov, E., Xiang, T., Gong, S.:
\newblock Semantic autoencoder for zero-shot learning.
\newblock In: Proceedings of the IEEE Conference on Computer Vision and Pattern
  Recognition. (2017)  3174--3183

\bibitem{lampert2009learning}
Lampert, C.H., Nickisch, H., Harmeling, S.:
\newblock Learning to detect unseen object classes by between-class attribute
  transfer.
\newblock In: 2009 IEEE Conference on Computer Vision and Pattern Recognition,
  IEEE (2009)  951--958

\bibitem{lampert2013attribute}
Lampert, C.H., Nickisch, H., Harmeling, S.:
\newblock Attribute-based classification for zero-shot visual object
  categorization.
\newblock IEEE transactions on pattern analysis and machine intelligence
  \textbf{36} (2013)  453--465

\bibitem{norouzi2013zero}
Norouzi, M., Mikolov, T., Bengio, S., Singer, Y., Shlens, J., Frome, A.,
  Corrado, G.S., Dean, J.:
\newblock Zero-shot learning by convex combination of semantic embeddings.
\newblock arXiv preprint arXiv:1312.5650 (2013)

\bibitem{rahman2018unified}
Rahman, S., Khan, S., Porikli, F.:
\newblock A unified approach for conventional zero-shot, generalized zero-shot,
  and few-shot learning.
\newblock IEEE Transactions on Image Processing \textbf{27} (2018)  5652--5667

\bibitem{xian2017zero}
Xian, Y., Schiele, B., Akata, Z.:
\newblock Zero-shot learning-the good, the bad and the ugly.
\newblock In: Proceedings of the IEEE Conference on Computer Vision and Pattern
  Recognition. (2017)  4582--4591

\bibitem{zhang2015zero}
Zhang, Z., Saligrama, V.:
\newblock Zero-shot learning via semantic similarity embedding.
\newblock In: Proceedings of the IEEE international conference on computer
  vision. (2015)  4166--4174

\bibitem{zhang2016zero2}
Zhang, Z., Saligrama, V.:
\newblock Zero-shot recognition via structured prediction.
\newblock In: European conference on computer vision, Springer (2016)  533--548

\bibitem{al2017automatic}
Al-Halah, Z., Stiefelhagen, R.:
\newblock Automatic discovery, association estimation and learning of semantic
  attributes for a thousand categories.
\newblock In: Proceedings of the IEEE Conference on Computer Vision and Pattern
  Recognition. (2017)  614--623

\bibitem{al2016recovering}
Al-Halah, Z., Tapaswi, M., Stiefelhagen, R.:
\newblock Recovering the missing link: Predicting class-attribute associations
  for unsupervised zero-shot learning.
\newblock In: Proceedings of the IEEE Conference on Computer Vision and Pattern
  Recognition. (2016)  5975--5984

\bibitem{xian2018zero}
Xian, Y., Lampert, C.H., Schiele, B., Akata, Z.:
\newblock Zero-shot learning—a comprehensive evaluation of the good, the bad
  and the ugly.
\newblock IEEE transactions on pattern analysis and machine intelligence
  \textbf{41} (2018)  2251--2265

\bibitem{zablocki2019context}
Zablocki, E., Bordes, P., Soulier, L., Piwowarski, B., Gallinari, P.:
\newblock Context-aware zero-shot learning for object recognition.
\newblock In: International Conference on Machine Learning, PMLR (2019)
  7292--7303

\bibitem{luo2019context}
Luo, R., Zhang, N., Han, B., Yang, L.:
\newblock Context-aware zero-shot recognition.
\newblock arXiv preprint arXiv:1904.09320 (2019)

\bibitem{krishna2017visual}
Krishna, R., Zhu, Y., Groth, O., Johnson, J., Hata, K., Kravitz, J., Chen, S.,
  Kalantidis, Y., Li, L.J., Shamma, D.A.,  et~al.:
\newblock Visual genome: Connecting language and vision using crowdsourced
  dense image annotations.
\newblock International journal of computer vision \textbf{123} (2017)  32--73

\bibitem{mishra2018generative}
Mishra, A., Krishna~Reddy, S., Mittal, A., Murthy, H.A.:
\newblock A generative model for zero shot learning using conditional
  variational autoencoders.
\newblock In: Proceedings of the IEEE Conference on Computer Vision and Pattern
  Recognition Workshops. (2018)  2188--2196

\bibitem{kumar2018generalized}
Kumar~Verma, V., Arora, G., Mishra, A., Rai, P.:
\newblock Generalized zero-shot learning via synthesized examples.
\newblock In: Proceedings of the IEEE conference on computer vision and pattern
  recognition. (2018)  4281--4289

\bibitem{verma2017simple}
Verma, V.K., Rai, P.:
\newblock A simple exponential family framework for zero-shot learning.
\newblock In: Joint European conference on machine learning and knowledge
  discovery in databases, Springer (2017)  792--808

\bibitem{verma2020meta}
Verma, V.K., Brahma, D., Rai, P.:
\newblock Meta-learning for generalized zero-shot learning.
\newblock In: AAAI. (2020)  6062--6069

\bibitem{yolo}
Redmon, J., Divvala, S., Girshick, R., Farhadi, A.:
\newblock You only look once: Unified, real-time object detection.
\newblock In: Proceedings of the IEEE conference on computer vision and pattern
  recognition. (2016)  779--788

\bibitem{ssd}
Liu, W., Anguelov, D., Erhan, D., Szegedy, C., Reed, S., Fu, C.Y., Berg, A.C.:
\newblock Ssd: Single shot multibox detector.
\newblock In: European conference on computer vision, Springer (2016)  21--37

\bibitem{retinaNet}
Lin, T.Y., Goyal, P., Girshick, R., He, K., Doll{\'a}r, P.:
\newblock Focal loss for dense object detection.
\newblock In: Proceedings of the IEEE international conference on computer
  vision. (2017)  2980--2988

\bibitem{fasterrcnn}
Ren, S., He, K., Girshick, R., Sun, J.:
\newblock Faster r-cnn: Towards real-time object detection with region proposal
  networks.
\newblock In: Advances in neural information processing systems. (2015)  91--99

\bibitem{rfcn}
Dai, J., Li, Y., He, K., Sun, J.:
\newblock R-fcn: Object detection via region-based fully convolutional
  networks.
\newblock In: Advances in neural information processing systems. (2016)
  379--387

\bibitem{maskrcnn}
He, K., Gkioxari, G., Doll{\'a}r, P., Girshick, R.:
\newblock Mask r-cnn.
\newblock In: Proceedings of the IEEE international conference on computer
  vision. (2017)  2961--2969

\bibitem{dcn}
Dai, J., Qi, H., Xiong, Y., Li, Y., Zhang, G., Hu, H., Wei, Y.:
\newblock Deformable convolutional networks.
\newblock In: Proceedings of the IEEE international conference on computer
  vision. (2017)  764--773

\bibitem{cornernet}
Law, H., Deng, J.:
\newblock Cornernet: Detecting objects as paired keypoints.
\newblock In: Proceedings of the European Conference on Computer Vision (ECCV).
  (2018)  734--750

\bibitem{centernet}
Duan, K., Bai, S., Xie, L., Qi, H., Huang, Q., Tian, Q.:
\newblock Centernet: Keypoint triplets for object detection.
\newblock In: Proceedings of the IEEE International Conference on Computer
  Vision. (2019)  6569--6578

\bibitem{fcos}
Tian, Z., Shen, C., Chen, H., He, T.:
\newblock Fcos: Fully convolutional one-stage object detection.
\newblock In: Proceedings of the IEEE International Conference on Computer
  Vision. (2019)  9627--9636

\bibitem{cai2018cascade}
Cai, Z., Vasconcelos, N.:
\newblock Cascade r-cnn: Delving into high quality object detection.
\newblock In: Proceedings of the IEEE conference on computer vision and pattern
  recognition. (2018)  6154--6162

\bibitem{redmon2017yolo9000}
Redmon, J., Farhadi, A.:
\newblock Yolo9000: better, faster, stronger.
\newblock In: Proceedings of the IEEE conference on computer vision and pattern
  recognition. (2017)  7263--7271

\bibitem{lin2014microsoft}
Lin, T.Y., Maire, M., Belongie, S., Hays, J., Perona, P., Ramanan, D.,
  Doll{\'a}r, P., Zitnick, C.L.:
\newblock Microsoft coco: Common objects in context.
\newblock In: European conference on computer vision, Springer (2014)  740--755

\bibitem{he2016deep}
He, K., Zhang, X., Ren, S., Sun, J.:
\newblock Deep residual learning for image recognition.
\newblock In: Proceedings of the IEEE conference on computer vision and pattern
  recognition. (2016)  770--778

\bibitem{lin2017feature}
Lin, T.Y., Doll{\'a}r, P., Girshick, R., He, K., Hariharan, B., Belongie, S.:
\newblock Feature pyramid networks for object detection.
\newblock In: Proceedings of the IEEE conference on computer vision and pattern
  recognition. (2017)  2117--2125

\bibitem{chua2009nus}
Chua, T.S., Tang, J., Hong, R., Li, H., Luo, Z., Zheng, Y.:
\newblock Nus-wide: a real-world web image database from national university of
  singapore.
\newblock In: Proceedings of the ACM international conference on image and
  video retrieval. (2009)  1--9

\bibitem{mikolov2013distributed}
Mikolov, T., Sutskever, I., Chen, K., Corrado, G.S., Dean, J.:
\newblock Distributed representations of words and phrases and their
  compositionality.
\newblock In: Advances in neural information processing systems. (2013)
  3111--3119

\bibitem{paszke2017automatic}
Paszke, A., Gross, S., Chintala, S., Chanan, G., Yang, E., DeVito, Z., Lin, Z.,
  Desmaison, A., Antiga, L., Lerer, A.:
\newblock Automatic differentiation in pytorch.
\newblock (2017)

\end{thebibliography}

%this would normally be the end of your paper, but you may also have an appendix
%within the given limit of number of pages
\end{document}